\documentclass[11pt,a4paper]{article}
\PassOptionsToPackage{hyphens}{url}\usepackage[hyperref]{emnlp2020}
\usepackage{times}
\usepackage{latexsym}

\usepackage{microtype}
\usepackage{arydshln}
\aclfinalcopy 

\usepackage{microtype}
\usepackage{amsmath}
\usepackage{graphicx}
\usepackage{amsthm}
\usepackage{multirow}
\usepackage{subfigure}
\usepackage{booktabs}
\usepackage{amssymb}
\usepackage{url}
\usepackage{url}
\usepackage{amsmath}

\usepackage{float}

\usepackage{cleveref}

\crefformat{section}{\S#2#1#3} 
\crefformat{subsection}{\S#2#1#3}
\crefformat{subsubsection}{\S#2#1#3}

\usepackage{color, colortbl}

\newcommand{\groc}[1]{GroC} 
\newcommand{\grocfull}[1]{the grounded compositional output model\xspace}

\title{Grounded Compositional Outputs for Adaptive Language Modeling }

\author{
Nikolaos Pappas$^{\clubsuit}$\quad 
Phoebe Mulcaire$^{\clubsuit}$\quad 
Noah A. Smith$^{\clubsuit}$$^{\diamondsuit}$\\
\\
  $^{\clubsuit}$Paul G.\ Allen School of Computer Science \& Engineering, University of Washington \\
   $^{\diamondsuit}$Allen Institute for Artificial Intelligence\\
   {\tt \{npappas,pmulcaire,nasmith@cs.washington.edu\} } 
}

\date{}
\begin{document}
\maketitle
\begin{abstract}
Language models have emerged as a central component across NLP, and a great deal of progress depends on the ability to cheaply adapt them (e.g., through finetuning) to new domains and tasks.  A language model's \emph{vocabulary}---typically selected before training and permanently fixed later---affects its size and is part of what makes it resistant to such adaptation. Prior work has used compositional input embeddings based on surface forms to ameliorate this issue. In this work, we go one step beyond and propose a fully compositional output embedding layer for language models, which is further grounded in information from a structured lexicon (WordNet), namely semantically related words and free-text definitions.  To our knowledge, the result is the first word-level language model with a size that does not depend on the training vocabulary.
We evaluate the model on conventional language modeling as well as challenging cross-domain settings with an open vocabulary, finding that it matches or outperforms previous state-of-the-art output embedding  methods and adaptation approaches.
Our analysis attributes the improvements to sample efficiency:  our model is more accurate for low-frequency words.
\end{abstract}

\section{Introduction}
 
Language models (LMs) are at the heart of natural language processing, especially following their recent success in the pretraining paradigm \citep[\emph{inter alia}]{dai15,peters-etal-2018-deep,devlin18,radford2019language}.  Continued advances in NLP rely on the adaptability of LMs to domains beyond their training data and to new domains and tasks, e.g.,~through domain adaptive pretraining followed by finetuning  \cite{gururangan-etal-2020-dont}.  Here, we focus on an important component of LMs, namely the \textbf{output vocabulary}---over which a LM's probability distribution over the ``next word'' (given the history) ranges---and investigate the impact of the type of its representation on the adaptability of neural LMs.

Today, LMs are typically trained with a closed output vocabulary derived from the training data; the vocabulary is not modified when the language model is adapted or deployed.  This makes large pretrained language models struggle with rare words,  despite being able to produce contextualized representations for them \cite{schick-schutze-2020-bertram}.  More importantly, this means a generative LM can never give nonzero probability to a specific word it did not see in training.  This is a longstanding challenge of language modeling \citep{jelinek1997statistical}, but it becomes especially important when we adapt to new domains and tasks.

One way to ``open up'' the vocabulary is to model sequences of bytes, characters, or ``wordpieces'' rather than the conventional word tokens \cite{sennrich15,Radford2018ImprovingLU,ponti-etal-2019-towards}.  While effective, this approach requires the LM to memorize subsequences if it is to treat them as words. These models appear to require greater network depth and show slower convergence than word-based alternatives \citep{cherry-etal-2018-revisiting,rami19}; the extra work comes at a cost. This is one of the reasons why the area of word-level language modeling is still very active \cite{baevski2018adaptive,sukhbaatar-etal-2019-adaptive,Khandelwal2020Generalization,press-etal-2020-improving}.

Interpolations between word- and character- or morphology-based LMs represent another class of solutions  \citep{mielke18,gerz-etal-2018-language,Ataman2020A}. These ``hybrid'' approaches  combine benefits from both model types. However, they introduce complexity which makes them potentially more difficult to train, maintain, and analyze. Notable for enabling adaptability are interpolated LMs based on  copy mechanisms \citep{merity2016pointer}, dynamic evaluation \cite{krause2017dynamic}, and neural caches \cite{grave2016improving,Grave17}; the last provides state-of-the-art adaptation performance and, unlike the rest, it does not require additional training. 

We propose a new word-level Grounded Compositional output LM (GroC)
that applies a compositional representation to the output vocabulary (Section~\ref{comp_output}).  Each word's output embedding is built from its surface character sequence and (if available) those of semantically related words and a free-text definition of from WordNet \citep{fellbaum98wordnet}.    This parameterization offers two chief advantages. First, GroC can assign probability to words not seen during training.  This means that a vocabulary different from the training vocabulary---e.g., one associated with a different text domain, crucial in adaptive settings---can be considered at inference time. Second, because there are no word type-specific parameters, the number of model parameters in GroC does not depend on the training vocabulary or its size.

 We evaluate GroC on language modeling with both fixed and open vocabularies in English. On standard language modeling (Section \ref{sec:closed_lm}) we observe that our model has superior perplexity and is more sample efficient than a variety of existing output embedding approaches, including the recent adaptive embedding of \citet{baevski2018adaptive}. 
 The open-vocabulary settings include a cross-domain setting and finetuning (Section~\ref{sec:cross}).  We find that GroC also outperforms strong interpolated baselines, including the unbounded neural cache model of \citet{Grave17} on ``near'' domains and performs competitively on ``far'' domains. 
 
Our analysis shows that our approach has improved sharing across words in the output vocabulary. We show experimentally that the perplexity gains are strongest for low-frequency words, implying improved \textbf{sample efficiency} relative to baselines: compositional output representations allow us to predict words from fewer training examples.

\section{Preliminaries on Language Modeling} \label{sec:lm}

Language models assign probability to sequences of tokens; the task is usually framed as learning the conditional probability distributions over individual tokens given their histories of tokens to the left \citep{bahl1983maximum}. 
Training requires a sequence of $T$ tokens $\boldsymbol{x} =  \langle x_1, \ldots, x_T\rangle$, each $x_t$ a member of a preselected vocabulary $\mathcal{V}$. We let $\mathbf{x}_t \in \{0,1\}^{|\mathcal{V}|}$ denote the one-hot encoding of $x_t$. 
The probability of the sequence $\boldsymbol{x}$ is factored using the chain rule of probability:
\vspace{-3mm}
\begin{align}
    p(\boldsymbol{x}) = \prod^{T}_{t=1} p(x_t \mid x_1, \ldots, x_{t-1}). \label{joint_prob}
\end{align}
To approximate this joint distribution, researchers have fit parametric families based on relative frequencies~\citep{bahl1983maximum,kneser1995improved,goodman2001bit}
and neural networks~\citep{bengio2003,mikolov10}.
Here, we focus on the latter due to their established effectiveness \cite{merity2017regularizing,baevski2018adaptive}. 
Tokens in this work correspond to words but they can also correspond to individual characters \cite{rami19} or byte pairs \cite{radford2019language}.

\subsection{Neural Language Models} \label{sec:nlms}

To make clear this paper's contributions, we describe neural language models by decomposing them into several abstract parts.

In most neural language models, the first layer of computation obtains an input embedding of each history word $x_j$ using a lookup function.  In our notation, this corresponds to selecting the word type's row in a fixed \textbf{input embedding matrix}, $\mathbf{E}^\mathit{in}$:  $\mathbf{x}_j^\top\mathbf{E}^\mathit{in}$, which we denote $\mathbf{e}^\mathit{in}_{x_j}$.
Importantly, however, input embeddings need not be lookups; for example, they can be built compositionally from the characters in the surface form of the word \citep{ling-etal-2015-finding}, an idea central to this work. 

Next, the history or ``prefix'' words ${\boldsymbol{x}_{<t}} = \langle x_1, \ldots, x_{t-1} \rangle$  is encoded into a fixed, $d$-dimensional vector $\mathbf{h}_{t-1}$ using a \textbf{prefix function} $f:\mathcal{V}^\ast\rightarrow \mathbb{R}^d$.  $f$ can be a recurrent or feedforward network; we will experiment with LSTMs \cite{hoch97} in Section \ref{sec:closed_lm}, but our method is agnostic to the prefix function design. In general, each history encoding is defined as
\begin{align}
    \mathbf{h}_{t-1} = f(\mathbf{e}^\mathit{in}_{x_1}, \ldots, \mathbf{e}^\mathit{in}_{x_{t-1}}). \label{eq:h}
\end{align}
Finally, the distribution over the next word (random variable $X_t$) is given by
\begin{align}
    p(X_t = x_t \mid \mathbf{h}_{t-1}) \propto \exp \left( \mathbf{E}^{\mathit{out}}\mathbf{h}_{t-1} + \mathbf{b}\right), \label{eq:lm}
\end{align}
where $\mathbf{E}^{\mathit{out}} \in \mathbb{R}^{|\mathcal{V}|\times d}$ is the \textbf{output embedding matrix} and $\mathbf{b} \in \mathbb{R}^{|\mathcal{V}|}$ is the bias vector (corresponding roughly to unigram log-frequencies of words in the vocabulary).  

The parameters of the model---including  all parameters of the prefix function $f$, $\mathbf{E}^{\mathit{in}}$, $\mathbf{E}^{\mathit{out}}$, and $\mathbf{b}$---are all chosen by maximizing the likelihood of the training sequence $\boldsymbol{x}$ under the model (Eq.~\ref{joint_prob}).  Note that, though we focus  on an autoregressive (left-to-right) language model objective, our analysis below is applicable to other language model pretraining objectives such as masked language modeling \cite{devlin18}  and replaced token detection \cite{Clark2020ELECTRA}. 
  
\subsection{Choice of Output Representations} \label{sec:choice}
\begin{figure}
  \centering
    \includegraphics[width=0.7\linewidth]{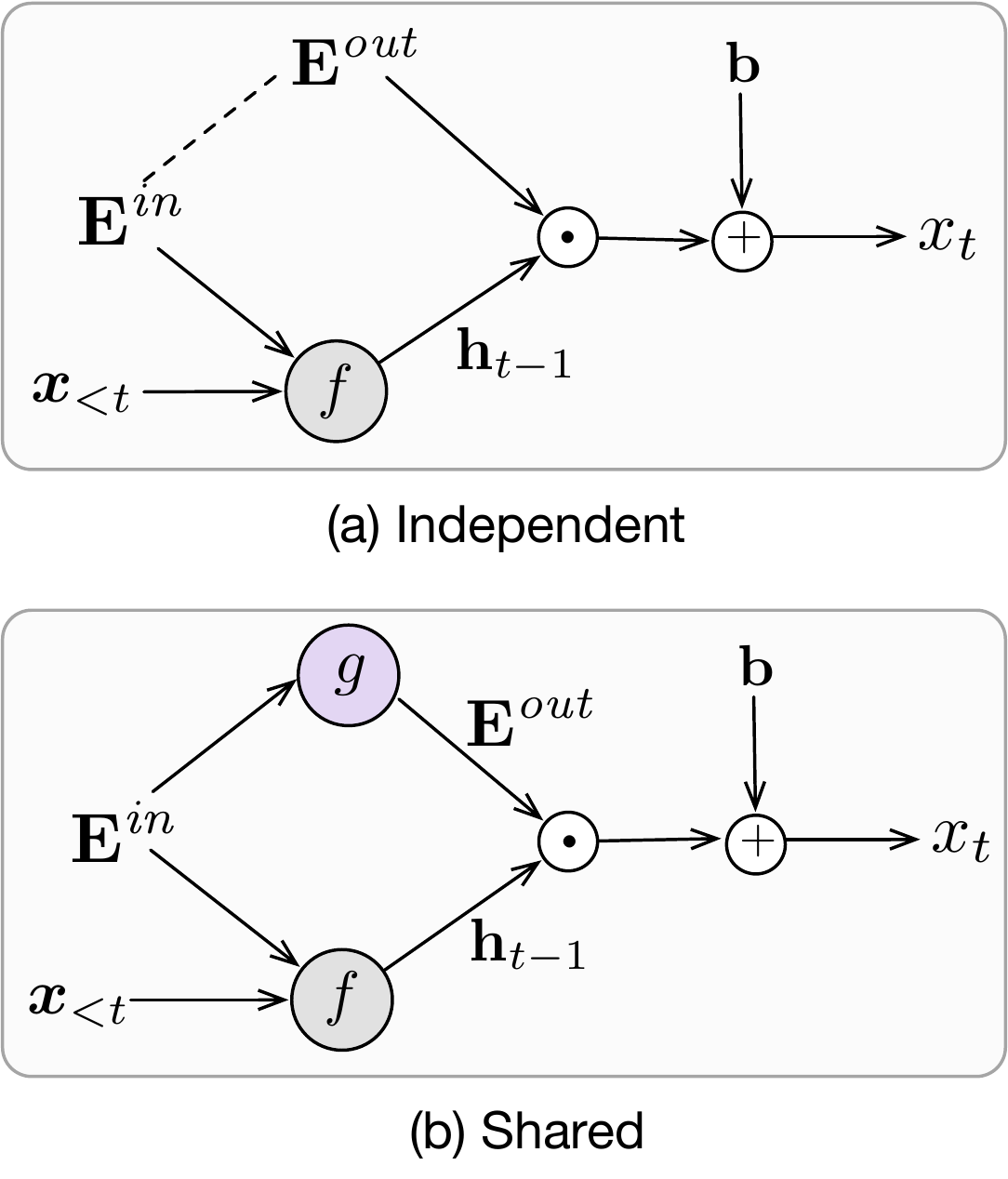} \\
    \vspace{-3mm}
   \caption{Existing output layer parameterizations.}
  \label{prior_output_emb}
  \vspace{-4mm}
\end{figure}

Above we assumed an output embedding matrix $\mathbf{E}^{\mathit{out}}$ that independently parameterizes each word in the vocabulary with a separate $d$-dimensional vector.  This approach requires $d\times|\mathcal{V}|$ parameters, leading to concerns about cost and overparameterization. 
Prior work addressed this issue by tying parameters between the input and output embedding matrices (i.e., $\mathbf{E}^{\mathit{out}} = \mathbf{E}^{\mathit{in}}$; \citealp{inan17,press17}). However, the parameters for each word are still independent from each other, as displayed in Figure \ref{prior_output_emb}(a).    

An alternative, also considered here, is to share output parameters across words as well as with the input embeddings. Specifically, this involves making the output embedding a function of the input embedding using a shared parameterization across words, $\mathbf{E}^{\mathit{out}} = g(\mathbf{E}^{\mathit{in}})$, as displayed in Figure \ref{prior_output_emb}(b).  For example,  \citet{gulordava18} used a linear transformation, while \citet{baevski2018adaptive} used a  linear transformation for each frequency bin to dedicate parameters to words proportional to their frequencies. \citet{pmlr-v97-pappas19a} used a deep residual transformation as $g$, demonstrating that shared parameterizations perform better than independent ones. The two latter studies also provided evidence that models with shared parameterization are more \textit{sample efficient} than independent parameterizations since they perform better on low-frequency words. 
 
\vspace{2mm} 

\noindent \textbf{Limitations}
We argue that dependence of a model's parameterization on the size of the vocabulary leads to several limitations shared by 
current word-level language models. First, the output embedding methods above have terms that scale with the vocabulary size, such as the lookup table for the input embedding or the bias vector, which is a concern for the parameterization of infrequent words. Second, handling of words \emph{unseen} in the training data leads us to the convention of uninformative ``out-of-vocabulary'' word types or linguistically na{\"\i}ve, data-driven vocabulary transformations that aggressively decompose words into smaller units \citep{sennrich15}.
Finally, when pretrained language models are adapted on a downstream task, they do not allow graceful modifications to the vocabulary as required by the task or its data domain. Decoupling the training vocabulary from the target vocabulary that a model can use during inference or finetuning will simplify sequential training and enable open vocabularies. 

Building on encouraging results with compositional  \emph{input} embeddings \citep{ling-etal-2015-finding,jozefowicz2016exploring,peters-etal-2018-deep}, we introduce a language model with shared compositional embeddings for {input} as well as for \emph{output} word representations. Further, we go beyond past work based on surface forms, making optional use of relations and natural language definitions from structured lexicons like WordNet \citep{fellbaum98wordnet}.  To our knowledge, this is the first word-level language model whose parameters do not depend on the vocabulary size and which is grounded to an external structured lexicon.
Our experiments show that our models are more sample efficient (Section \ref{sec:closed_lm}) on closed vocabularies and perform  competitively on cross-domain settings (Section \ref{sec:cross}).
 
\begin{figure*}[h]
\hspace{2.5mm}\includegraphics[width=15.5cm]{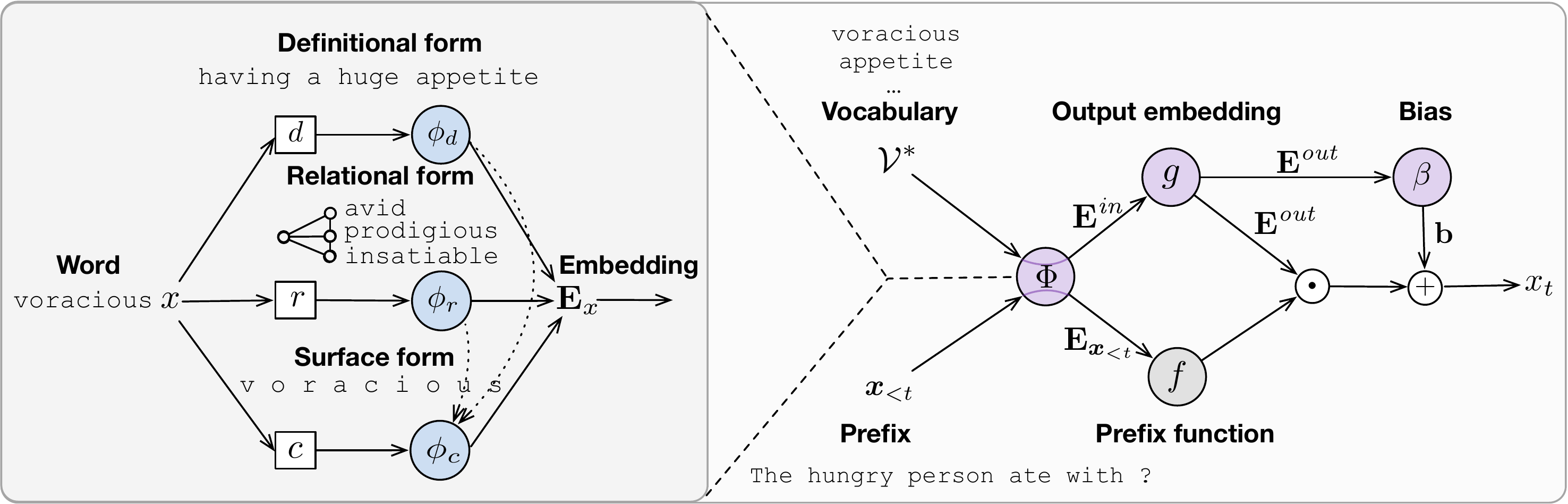}
\caption{
Grounded compositional output language modeling. (\textit{Left}) The compositional input embedding is grounded in surface, relational, and definitional word forms from an external structured lexicon. (\textit{Right}) The encoded prefix words are given as input to the prefix function and the words in an arbitrary vocabulary are given as input to the output embedding function and the bias function to predict the next word.}
\label{diagram}
\vspace{-3mm}
\end{figure*}

\section{GroC: Grounded Compositional Output Language Models} 
\label{comp_output}

We present our grounded compositional output language model (Figure~\ref{diagram}).\footnote{Code: \url{https://github.com/Noahs-ARK/groc}}
Following the decomposition of neural language models in Section~\ref{sec:lm} (Equations~\ref{eq:h}--\ref{eq:lm}), we consider each part of the model in turn:  input embeddings (Section~\ref{sec:compositional}), output embeddings (Section~\ref{sec:output_emb}),   and bias (Section~\ref{sec:bias_estimator}). 
As noted above, our approach is agnostic to the training vocabulary ($\mathcal{V}$) and to the prefix encoder ($f$) that has been the focus of most innovations in neural language model design.

\subsection{Compositional Input Embeddings} \label{sec:compositional}

We build on the compositional model of  \citet{ling-etal-2015-finding}, which encodes a word using its surface string (i.e., character sequence), adding two more sources of information.  \citet{peters-etal-2019-knowledge} enhanced word representations with information from external relational knowledge bases, specifically for words that refer to entitites.  Like them, we use a structured lexicon (WordNet); we encode every word in the lexicon using its neighbors.
The second follows \citet{bahdanau17}, who used \textbf{definitions} to represent out-of-vocabulary words; we encode definitions for all words (regardless of training-set frequency).

We begin by replacing the matrix $\mathbf{E}^{\mathit{in}} \in \mathbb{R}^{|\mathcal{V}| \times d}$ with a neural network that defines a word's embedding compositionally from its surface form, its position relative to other words in a structured lexicon, 
and a natural language definition. For each word $x$, we refer to these, respectively, as the word type's surface embedding $\mathbf{c}_x$, relational embedding $\mathbf{r}_x$, and definitional embedding $\mathbf{d}_x$. We assume each has a dimensionality of $d$.  The last two are optional (if missing, they are set to zero), and we redefine $\mathbf{e}_{x}$ as the concatenation of the three, namely $\mathbf{e}_{x} = \langle \mathbf{c}_x,   \mathbf{r}_x, \mathbf{d}_x\rangle$.  For $\mathbf{r}_x$ and $\mathbf{d}_x$, we used the structured relations (synonyms and hyponyms) and free-text definitions in WordNet \cite{fellbaum98wordnet}.  

In this study, we focus on simple, computationally efficient options for the three encoders.  A word $x$'s character sequence is encoded as surface encoding $\mathbf{c}_x$ using a convolutional network followed by a highway network
\citep{jozefowicz2016exploring,peters-etal-2018-deep}.  Its relational encoding $\mathbf{r}_x$ is given by an average of $\mathbf{c}_{x'}$ across WordNet synonyms and hyponyms $x'$. The definitional encoding of $x$, $\mathbf{d}_x$, we similarly take an average of the surface encodings $\mathbf{c}_{x'}$ over words $x'$ appearing in the definition. For computational efficiency, we set a maximum limit to the number of words to be used for both relations and definitions (see Appendix \ref{sec:model_config}).   

If a word's information is not in WordNet, we set $\mathbf{r}_x$ and/or $\mathbf{d}_x$ to $\boldsymbol{0}$.
In future work, additional encodings could be appended, such as contextualized examples \cite{Khandelwal2020Generalization}.

A notable property of these input embeddings is that their parameter count does not depend on the vocabulary size $|\mathcal{V}|$.  Further, the vocabulary used in training need not be identical to the one used during finetuning, evaluation, or deployment. For example, during training we can use the full vocabulary combined with a softmax approximation method (e.g., \citealp{pmlr-v70-grave17a}), or by dynamically narrowing the choice of $x_t$ based on its history using co-occurrence statistics \cite{lHostisGA16}.  During finetuning or evaluation, one can use the same vocabulary (required for traditional perplexity evaluations) or a different one chosen statically or dynamically, since any word's input embedding can be calculated compositionally.

\subsection{Compositional Output Embeddings}
\label{sec:output_emb}

One straightforward option for vocabulary size-independent output embeddings is to reuse the compositional input embeddings from Section~\ref{sec:compositional}, along the lines of \citet{press17}.  Concretely, at timestep $t$, we take the set $\mathcal{V}'_t$ of output word types allowed, embed each word type ${v\in\mathcal{V}'_t}$ as in Section~\ref{sec:compositional}, and stack these into a matrix $\mathbf{E}^{\mathit{in}}_{t}$  which serves directly as $\mathbf{E}^{\mathit{out}}$. 

Though these compositional representations do enable extensive sharing across the vocabulary, we suspect that the features they capture may require additional processing before capturing ``output'' distributional similarity, especially when another domain 
is the real target use case for the language model. This follows prior work discussed in Section \ref{sec:choice}, which showed that making the output embedding a function of the input embeddings with shared parameters improves over simple tying.\footnote{Note that the input embeddings are passed through the prefix encoder $f$, which uses additional parameters to create the hidden state $\mathbf{h}_{t-1}$.} 

We therefore adopt a depth-$k$ residual network for the output embedding function $g$ (from  Section \ref{sec:choice}) that consists of a feedforward function $g_i$ at each layer $j$ with $d$-dimensions each and apply it to the input embedding at timestep $t$:
\begin{align}
    \forall j: 1 \leq j \leq k, {\mathbf{E}^{\mathit{out}}_t}^{(j)} & = g_j\left( {\mathbf{E}^{\mathit{out}}_t}^{(j-1)}\right) + \mathbf{E}^{\mathit{in}}_t \nonumber \\
   {\mathbf{E}^{\mathit{out}}_t}^{(0)} & = \mathbf{E}^{\mathit{in}}_t.
\end{align}
Hence, we use ${\mathbf{E}^{\mathit{out}}_t}^{(k)}$ as the output embedding at timestep $t$. To avoid overfitting, we  apply variational dropout in between the layers, following \citet{pmlr-v97-pappas19a}. In contrast to that work, our resulting output embeddings are compositional.
 The depth $k$ and the dropout rate are hyperparameters to be tuned on development data. The number of parameters is proportional to $k$ times the number of parameters in the feedforward network ($O(d^2)$); it does not depend on the vocabulary size.  

\subsection{Bias}\label{sec:bias_estimator}
\label{bias_estimator}

In conventional language models, each word in the vocabulary is assigned a bias parameter that roughly captures its log-frequency under a unigram distribution.  This is the last part of a neural language model whose parameters depend on the vocabulary size.  Instead of a dedicated, independent bias parameter for each word $v \in \mathcal{V}$, we define
\vspace{-1mm}
\begin{align}
    b_v &= \sigma\left(\mathbf{w} \cdot \mathbf{e}^{\mathit{out}}_v + a\right),
\end{align}
where $\sigma$ is the activation function and we introduce parameters $\mathbf{w} \in \mathbb{R}^d$ and $a \in \mathbb{R}$.  The bias values $b_v$ are stacked to form $\mathbf{b}$ and used in Equation~\ref{eq:lm}.

\subsection{Training}\label{sec:training}
Since all components are differentiable with respect to their parameters, the entire model can be trained to maximize training-data likelihood as described earlier (Section \ref{sec:nlms}).
Parameters include:
\begin{itemize}
    \itemsep -0.1em
    \item Input character embeddings, the convolutional network for $\mathbf{c}_\ast$, and $3d^2$ parameters for projection (Section~\ref{sec:compositional});
    \item Output embedding transformation, including the depth-$k$ feedforward network for output embeddings (Section~\ref{sec:output_emb}) and the bias parameters (Section~\ref{sec:bias_estimator}); and
    \item Prefix encoder $f$, an orthogonal design choice to our method (an LSTM in our experiments).
\end{itemize}

The model size can be adjusted by changing output embedding hyperparameters to fit a given memory requirement  — this is the same as any other neural network.  Note that despite our vocabulary-size independent parameterization,  we still need to process all the words in the supplied vocabulary leading to increased training times despite the model's sample efficiency.
This can be prohibitive for very large vocabularies ($\geq 100K$), where we recommend using softmax approximation methods and making sparse updates of the output embedding parameters (see Appendix 1.3). 
During inference, $\textbf{E}^{\textit{out}}$ can be 
cached for fast access; there is no need to execute a forward pass more than once.

\section{Conventional Language Modeling} 
\label{sec:closed_lm}

We first establish the performance of \groc~ in the conventional closed-vocabulary setting, considering two datasets. We consider out-of-sample generalization (measured by test-set perplexity) and also analyze fit across the vocabulary by frequency bin.

\subsection{Experimental Setup}
\label{sec:conventional_setup}

\paragraph{Datasets.} We evaluate our methods on two English datasets: 
\texttt{penn} \cite{ptb93} and \texttt{wikitext2} \cite{merity2016pointer}. 
We report test perplexity using the provided  training/dev./test splits (see details in Appendix \ref{sup:dev_scores}). 
Table~\ref{tab:lm_data} also quantifies the percentage of each dataset's vocabulary that is covered by WordNet (used to derive relational and definitional encodings).

\begin{table}[htp]
    \centering \small
  \def\arraystretch{1.2}\tabcolsep=6.4pt    
    \begin{tabular}{llrrrr}
    \toprule
        \textbf{Dataset} & genre & $|\mathcal{V}|$ & \# tokens & \multicolumn{2}{c}{WNet cov.}\\ \midrule
         \texttt{penn}& news & 10K & 929K & 78 & 86 \\
         \texttt{wikitext2} & Wikip. & 33K & 2M & 73 & 76 \\ 
         \bottomrule
    \end{tabular}
    \caption{Language modeling dataset statistics.  
    \label{tab:lm_data}}
    \vspace{-5mm}
\end{table}

\paragraph{Models.} All of the models compared use the same prefix encoder:   a vanilla recurrent neural network based on the implementation by \citet{merity2016pointer} with 2 layers and 1024 LSTM units, regularized with hidden unit dropout of $0.65$ along the lines of \citet{Grave17}. More details are given in Appendix \ref{sec:training_speed}. The following output embedding approaches are compared: 
\begin{itemize}
    \itemsep 0em
    \item \textbf{Lookup table}: trains a full output embedding lookup table that corresponds to the vocabulary as defined in Eq.~\ref{eq:lm}.
    
    \item \textbf{Convolutional} \cite{jozefowicz2016exploring}: an alternative to a lookup table that uses a character-level convolutional neural network followed by a highway network plus a linear ``correction'' for each vocabulary element to represent the outputs.\footnote{Note that we chose not to use a linear ``correction'' with \groc~ since it deviates from our goal of having a vocabulary-independent parameterization, but it could be applied to \groc~ in the future for additional improvements.} 
    
    \item \textbf{Tied} \cite{press17}: avoids training separate input and output embedding matrices by tying their parameters. This is a common technique that mitigates the overparameterization issue of the lookup table. 
    \begin{figure*}
\vspace{-5mm}
  \centering 
  \includegraphics[width=15.2cm]{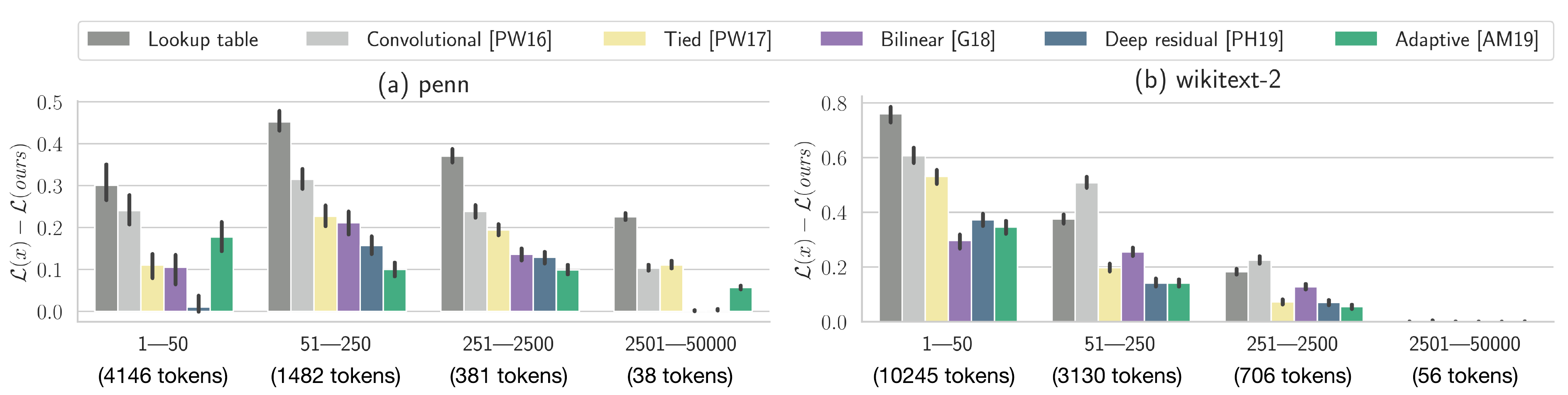}   
\vspace{-3mm}
      \caption{ 
      Median loss difference between each baseline and \groc~ over different word frequency intervals on   \texttt{penn} (a) and \texttt{wikitext2} (b). The biggest differences are mostly observed on words with low training frequencies.
  \label{fig:cediff}}
  \vspace{-3mm}
\end{figure*} 
    
    \item \textbf{Bilinear} \cite{gulordava18}: performs a simple linear transformation of the input embedding to produce the output embedding that effectively shares parameters across outputs. 
    
    \item \textbf{Deep residual} \cite{pmlr-v97-pappas19a}: performs a deep residual transformation of the input embedding with variational dropout in between its layers, which is more expressive than the bilinear one. 
    
    \item \textbf{Adaptive} \cite{baevski2018adaptive}: uses a bilinear transformation of the  input and output embedding with parameters proportional to the word frequencies, to assign more capacity to frequent words and less capacity to infrequent ones. This is considered to be a state-of-the-art output embedding method.  
    \item \textbf{\groc~} (ours): the grounded compositional output embedding 
   described in Section~\ref{sec:output_emb}.
\end{itemize}

\begin{table}
  \centering \small
  \def\arraystretch{1.1}\tabcolsep=6pt    
  \begin{tabular}{l rr rr }
    \toprule
      & \multicolumn{2}{c}{\texttt{penn}} & \multicolumn{2}{c}{\texttt{wikitext2}} 
      \\
    \textbf{Output embedding} & {$|\Theta|$ }  & {test}  & {$|\Theta|$}  & {test} 
    \\
    \toprule
    Lookup table                                       & 13M &  90.8 & 23M  & 108.3
    \\
    Convolutional  [\textcolor{black}{J16}]             & 13M & 101.6 & 23M & 116.6  
    \\
    Tied [\textcolor{black}{PW17}]                      & 10M  & 86.2 & 15M & 97.3
    \\
    Bilinear [\textcolor{black}{G18}]                   & 10M & 83.7 & 15M & 95.9
    \\
    Deep residual [\textcolor{black}{PH19}]             & 10M & 80.5 & 15M & 94.7
    \\
    Adaptive [\textcolor{black}{AM19}]                  &  8M & 79.3 & 9M  & 90.7
    \\\midrule
    \groc~ (ours)               &  9M  & {69.5} & 9M  & {82.5}
    \\ 
    \bottomrule
    
  \end{tabular}
  \vspace{-1mm}
  \caption{Perplexity scores on conventional language modeling benchmarks with closed vocabulary. $|\Theta|$ denotes the total number of model parameters.
  \label{tab:lm_closed}}
  \vspace{-5mm}
\end{table}

For fair comparison, we apply variational dropout to all output embeddings.  Hyperparameter selection of dropout rates, output network depth and activation, linear ``correction,'' and adaptive frequency cutoffs was conducted by grid search on validation data.  Details are given in Appendix \ref{sec:lm_hyper}.

\subsection{Results}
Table~\ref{tab:lm_closed} reports perplexities achieved by all seven models.  
The main finding is that \groc~ achieves lower perplexity than the previous models, on both datasets.  Note that \groc~ outperforms the state-of-the-art output embedding method of \citet{baevski2018adaptive}; specifically, by $-9.8$ and $-8.2$ points on \texttt{penn} and \texttt{wikitext-2}, respectively. The difference with the other methods is even larger.
We also confirm the findings of  \citet{pmlr-v97-pappas19a}, that output parameter sharing methods outperform tied output embedding and the lookup table, and, of \citet{jozefowicz2016exploring}, that convolutional output embeddings lag behind full softmax (lookup table). Notably, \groc~ outperforms the best reported scores by \citet{merity2016pointer} and \citet{Grave17} on \texttt{penn}, using about 11M fewer parameters and a similar prefix network to the latter. See Appendix \ref{sec:sota} for a more detailed comparison with state-of-the-art models of similar size.

Nevertheless, \groc~ is about $1.3\times$ slower than the convolutional method on \texttt{penn}; with sparse updates ($p>0.3$) we can make it $2.1\times$ faster than that method,  which is comparable to the speed of the  bilinear method, while maintaining a perplexity improvement of $-26$ points  (see detailed speed comparisons in Table \ref{tab:train_speed} in Appendix \ref{sec:training_speed}).

\subsection{Analysis} \label{sec:analysis}

The experiment above establishes that our approach achieves improved perplexity relative to alternative output embeddings.  We next decompose its performance in various ways to understand why.

\paragraph{Word frequency effects.} 
We conjecture that \groc~'s main benefit comes from words that are rare in the training data, since the core contribution is to share representations across the vocabulary.  To evaluate this hypothesis, we consider the difference in test loss (cross entropy) between \groc~ and a baseline model, following \citet{baevski2018adaptive} but computing the median instead of the average to reduce the effect of outliers.  
We decompose this score by data frequency bins (e.g., words occuring 1--50 times in the training dataset). 
Figure~\ref{fig:cediff} is displayed for the \texttt{penn} and \texttt{wikitext2} datasets. 
The trend we observe is that \groc~ has the greatest relative benefit for words in lower frequency bins, compared to each model.  The lowest-frequency bin on \texttt{penn} deviates from this pattern,  which we take as an indication that generalizing to infrequent words with only 1M training tokens and a small 10K vocabulary is inherently challenging.

\paragraph{Ablations.}

To assess the contributions of GroC's components, we performed ablation tests on \texttt{penn} and \texttt{wikitext2}
(Table \ref{tab:ablation}).
These include removing relational and/or definitional forms, either with or without a deep residual output network.
For fairness, we tune the hyperparameters of the ablated model  variants as above. 
Overall, removing the relational and definitional forms from the main model with or without output network on top increases the perplexity. The largest drop in perplexity happens when we remove both forms, which highlights their notable contribution to the full model. 
Lastly, the results on \texttt{wikitext2} highlight the importance of capturing the output similarity with an output network (out) for datasets with a larger vocabulary as opposed to merely reusing the grounded compositional embeddings as output embeddings. 
\begin{table}[ht]
  \centering \small
  \def\arraystretch{1.1}\tabcolsep=9pt
  \begin{tabular}{lcccc}
    \toprule
     &  \multicolumn{2}{c}{\texttt{penn}} & \multicolumn{2}{c}{\texttt{wikitext2}}\\
    \textbf{Model} &  dev. & test  &   dev. & test\\
    \toprule 
    \groc~ $+$ out       & 75.0  & 71.4  - &   {87.0}  & {82.5} -\\ 
     $-$ relations                                   & 77.1  &  72.7 \textcolor{red}{$\uparrow$}  &  90.2  & 85.3 \textcolor{red}{$\uparrow$} \\
     $-$ definitions                                 & 75.6  &  72.0  \textcolor{red}{$\uparrow$}  &  88.6  & 84.3  \textcolor{red}{$\uparrow$} \\ 
     $-$ both                                       & 79.8  &  75.8 \textcolor{red}{$\uparrow$}  & 94.3 & 89.8 \textcolor{red}{$\uparrow$} \\ 
     \hline
     \groc~                                  & {72.5}  & {69.5}  - & 88.7 & 84.1 -   \\ 
    $-$ relations                                         & 74.2  & 70.8  \textcolor{red}{$\uparrow$} & 93.0 & 88.0  \textcolor{red}{$\uparrow$}\\
     $-$ definitions                                          & 74.4  & 71.1  \textcolor{red}{$\uparrow$}  & 87.6 & 83.1 \textcolor{blue}{$\downarrow$} \\ 
    $-$ both                                              & 76.3  & 73.2  \textcolor{red}{$\uparrow$}&  94.5  & 89.5  \textcolor{red}{$\uparrow$}  \\ 
    \bottomrule
  \end{tabular}
  \vspace{-2mm}
  \caption{Ablated model variants on \texttt{penn} and \texttt{wikitext-2}.
  \textit{out}: the deep residual output network. 
  \label{tab:ablation}}
  \vspace{-11mm}
\end{table}

\vspace{3mm}

\paragraph{Lexicon coverage.}  To measure the effect of lexicon coverage on model performance in a controlled setting, we artificially remove words from WordNet, making them unavailable for relational and definitional encodings.  In this experiment, we consider the \texttt{penn} dataset, where WordNet's coverage over the (relatively small) vocabulary is highest to begin with. 
Table~\ref{tab:coverage} shows the resulting test perplexity of a pretrained model (inference) and a model trained from scratch (train) when such controlled manipulation is applied to them from 0\% up to the maximum of 82\% coverage (Table~\ref{tab:lm_data}) . Note that we treat relational independently of definitional forms since they are not always co-present. Overall, the results indicate that the model is sensitive to changes in the forms of words that have been seen  during training but it is robust to changes if it is trained from scratch. 
\begin{table}[htp]
  \centering \small
  \def\arraystretch{1.2}\tabcolsep=3.2pt
  \begin{tabular}{cccccccc}
      \toprule
        \textbf{Coverage} & surf.  & $0\%$ & $16\%$ & $32\%$ & $48\%$ & $64\%$ & $82\%$ \\ 
     \toprule
      inference       & 73.1 & 187.8  & 159.3 & 128.5 & 102.6 & 83.5 & 69.5 \\
        train         &  -- & 72.4   & 70.0  & 70.4  & 69.6  & 70.7 & 69.5 \\
     \bottomrule
  \end{tabular}
  \vspace{-2mm}
  \caption{External lexicon coverage effect on the perplexity of \groc~ on the $\texttt{penn}$ test
  set. \textit{surf.}: model with surface forms only from Table \ref{tab:ablation}, last row.}
  \label{tab:coverage}
  \vspace{-4mm}
\end{table}In the next section, we investigate what happens when we add forms for words which the model has never seen before.

\begin{table*}
  \centering \small
  \def\arraystretch{1.1}\tabcolsep=8pt    
  \begin{tabular}{p{0pt} p{0pt} l c  c c c c | c  c }
    \toprule
   & & & & \multicolumn{4}{c}{{{near} domains}} & \multicolumn{2}{c}{{{far} domains}} \\ 
   & & \textbf{Model} & {\hspace{3mm}  \texttt{2007}} $\mathbf{\rightarrow}$     & {  \texttt{2008}}
      & { \texttt{2009}} &
       {  \texttt{2010}} & 
        {  \texttt{2011}} & 
        {   \texttt{Web}} &
      {   \texttt{Wiki}} \\ 
    \toprule
      \multicolumn{10}{c}{Zero-resource setting} \\\midrule 
   \parbox[t]{0mm}{\multirow{4}{*}{\rotatebox[origin=c]{90}{ { \scriptsize \bf \textcolor{darkblue}{Grave et al.}}   }}} & \parbox[t]{0mm}{\multirow{4}{*}{\rotatebox[origin=c]{90}{ { \scriptsize \bf \textcolor{darkblue}{(2017a)}}   }}} &     Base              & 220.9 &237.6 & 256.2 & 259.7 & 268.8 & 689.3 &  1003.2\\  
  &  & Base + unigram         & 220.3   &235.9 & 252.6 &256.1 &  264.3 & 581.1 & 609.4  \\
  & & Base + local cache    & 218.9 & 234.5&  250.5  &  256.2 &  265.2 &   593.4 & 316.5 \\  
  & &  Base + unbounded cache    & 166.5 & 191.4 & 202.6 & 204.8 & 214.3 & 383.4 & 337.4 \\
    \hline \midrule 
 \parbox[t]{0mm}{\multirow{2}{*}{\rotatebox[origin=c]{90}{ { \scriptsize \textbf{Baseline}}   }}} & &  Tied         & 184.3 & 199.8 &  217.3 & 221.6 &  229.9 & 660.6 & 841.1 \\
 &&  Tied + unigram [\textcolor{black}{G17}]    & 187.8 & 203.6 &  221.5 & 225.9 &  234.3 & 577.5 & 819.7 \\
  & &  Tied + local cache [\textcolor{black}{G17}]   & 181.8 & 196.5 &  212.0 & 217.7 &  225.9 & 501.7 & \bf 406.9 \\ \midrule
 \parbox[t]{0mm}{\multirow{3}{*}{\rotatebox[origin=c]{90}{ { \scriptsize \bf Ours}   }}}   & &  \groc~                                    & 158.6  &  171.0 & 186.7 & 192.5 & 200.4 & 637.9  &  753.9  \\
  & &  \groc~ + unigram [\textcolor{black}{G17}]    & 155.2 &  167.3  & 183.1 & 189.5 & 196.4  & 533.6  & 689.2 \\
  & &  \groc~ + local cache [\textcolor{black}{G17}]     & \bf 152.6  &  \bf 164.1 & \bf 179.0  & \bf 185.1 & \bf 192.3   &\bf 493.0  &  408.8 \\
    \midrule
        \multicolumn{10}{c}{Low-resource setting} \\\hline
    &  &  Tied + finetuning    & -- & 172.8  &  177.9  & 180.7 & 185.4 & \bf 212.7 & 242.6  \\  
  & & \groc~ + finetuning       &  -- &  \bf153.7  & \bf162.2  &\bf 167.0 & \bf 170.6 & 239.5  &\bf  216.9  \\
    \bottomrule 
  \end{tabular}
  \vspace{-2mm}
  \caption{Results on \textit{near} and \textit{far} cross-domain language modeling with an open vocabulary with a zero-resource or a low-resource setting. Top four rows display scores from  \citet{Grave17}, while the next three are from our re-implementation with a stronger base model. Boldface marks the best perplexity on each test set. 
  }
  \label{tab:lm_cross}
  \vspace{-3mm}
\end{table*}

\section{Cross-Domain Language Modeling}
\label{sec:cross}
To demonstrate our model's ability to generalize beyond its training data, we evaluate it across domains with an open vocabulary, in two settings: zero-resource, where it is first trained on one domain and then tested on a new target domain,  
and low-resource, in which the model is further exposed to training data in the new domain.

\subsection{Experimental Setup}

\paragraph{Data.} 
Following \citet{Grave17}, we create English datasets from News Crawl \cite{bojar2014findings}, Common Crawl,\footnote{We used the version from WMT 2014 \cite{bojar2014findings}.} and WikiText-103 \cite{merity2016pointer}. Dataset statistics are given in Table \ref{tab:cd_stats}. All models are trained on 2M tokens from the \texttt{2007} dataset and evaluated on 10M tokens; finetuning is done on an additional 2M tokens  from the target domain.  We consider the domain of the \texttt{2008-2011} datasets to be similar (``near'') to that of the training set, \texttt{2007}, as they contain news from different time periods. In comparison, \texttt{web} and \texttt{wiki} are more different (``far'') from \texttt{2007}.

\begin{table}
  \centering \small
  \def\arraystretch{1.2}\tabcolsep=3.5pt    
  \begin{tabular}{ c c c c  c }
  \toprule
      \textbf{Dataset} & source & train $|V|$  & test $|V|$ & {OOV\%} \\
  \toprule
    \texttt{2007} & \multirow{6}{*}{News Crawl}  & 81K & 188K & 2.0  \\
    \texttt{2008} & & 82K & 197K & 2.3 \\
    \texttt{2009} & & 81K & 195K & 2.5 \\
    \texttt{2010} & & 78K & 181K & 2.4  \\
    \texttt{2011} & & 80K & 184K & 2.5   \\
  \hline
  \texttt{web} & Common Crawl &  75K & 174K & 5.8 \\
  \hline
  \texttt{wiki} & WikiText-103&   67K & 109K & 5.4 \\
  \bottomrule
  \end{tabular}
  \vspace{-1mm}
  \caption{Dataset statistics for cross-domain experiments. OOV\% gives the percentage of tokens in the test set not present in the \texttt{2007} train vocabulary.}
  \label{tab:cd_stats}
  \vspace{-5mm}
\end{table}

\paragraph{Models.}
We compare GroC to the tied output embedding model described in Section \ref{sec:conventional_setup} when combined with the following adaptation methods:
\begin{itemize}
    \itemsep 0em
    \item \textbf{Unigram}: we interpolate the model's distribution with a unigram cache, which assigns probabilities based on the counts of words in the test data observed so far during evaluation.
    \item \textbf{Local cache}: we interpolate the model's distribution with a neural cache \cite{grave2016improving}, which assigns probabilities based on the similarity of the current hidden state to previous hidden states during evaluation.
    \item \textbf{Finetuning}: the model is finetuned on 2M tokens from the target domain.
\end{itemize}
(We also compare to the reported unbounded cache results from \citealp{Grave17}.) Cache models provide effective adaptation without training by using recent history to develop an auxiliary distribution during evaluation, informing predictions of unseen or rarely-seen words. 
However, as \groc~ already assigns non-negligible weight to new words not seen prior to evaluation, the cache has less effect by default, even if its predictions are more accurate, an effect we observed in validation.
To address this, we down-weighted the model's predictions for new words prior to cache interpolation by 0.1. For finetuning, both tied and \groc~ models were trained for an additional 3 epochs on the target domain, allowing them to adapt to the new domain. See Appendix \ref{sec:app_cross_hyper} for hyperparameter details.

\paragraph{Vocabulary setting.}
For a fair comparison, all models are evaluated on the union of the training and test vocabularies. 
Tied models are interpolated with the uniform distribution at test time to prevent infinite perplexities on unseen words, prior to cache interpolation if applicable. 
Words present in the finetuning data but not in the original training data are given random embeddings prior to finetuning. 

\subsection{Results} \label{sec:open_lm}
The results for the cross-domain experiments are shown in Table
\ref{tab:lm_cross}. Standalone \groc~ improves perplexity relative to the tied model in every domain by up to $-30$ points, including the local neural cache and the unbounded neural cache model in the near-domain, even when the former is applied to our own stronger tied-embedding baseline model. 
In addition, finetuned \groc~ outperforms all previous baselines by a wide margin including the unbounded cache by about $-40$ and $-132$ points on near and far domains, respectively.
Here, \groc~ outperforms the finetuned tied model  by up to $-25$ points except in \texttt{web} domain, and reaches lower validation scores with fewer iterations in 5 out of 6 domains (see Appendix \ref{sec:app_cross_finetuning}).  
For the \texttt{web} domain, caches and finetuning are more effective than in any other domain, indicating unique domain dynamics worthy of further study.

\section{Conclusion}

We proposed an adaptive language model based on grounded compositional outputs. We demonstrated that it reduces the number of parameters and increases sample efficiency, outperforming strong output embedding methods and adaptation baselines on both in-domain and open-vocabulary settings respectively. In principle, our results should be applicable to word-piece language models which are currently based on lookup tables to improve their sample efficiency and compactness. In future work, it would be interesting to investigate to what extent pretrained language models benefit from \groc~ on such zero-resource or low-resource adaptation settings. This work indicates several other future directions for language modeling in low-resource domains: extension to other languages,  scaling training to even larger vocabularies, and applying \groc~ in a large pretraining setting to expand its zero-shot generalization.

\section*{Acknowledgments}
We thank Sofia Serrano, Ofir Press, Yizhong Wang, Mourad Heddaya, Kelvin Luu, and the anonymous reviewers for their helpful feedback. Nikolaos Pappas was supported by the Swiss National Science Foundation under the project UNISON, grant number P400P2\_183911.  This work was supported in part by US NSF grant 1562364 and the Office of Naval Research under MURI grant N00014-18-1-2670.

\bibliography{references}
\bibliographystyle{acl_natbib}
 
 \clearpage
 \appendix 
  




\section{Supplementary Material for ``Grounded Compositional Outputs for Adaptive Language Modeling''}
We report here the computer infrastructure, experimental details, including  hyperparameter bounds, hyperparameter optimal values, training speed, development scores, for both of the experiments where applicable. We also provide a comparison with state of the art by taking into account the number of model parameters and guide the reader through the  replication effort we did to reproduce the neural cache by \citet{Grave17}.

\section{Conventional Language Modeling}\label{sec:conlm_details}
For the experiments with a closed vocabulary on \texttt{penn}\footnote{\url{www.fit.vutbr.cz/~imikolov/rnnlm/simple-examples.tgz}} and \texttt{wikitext-2},\footnote{\url{s3.amazonaws.com/research.metamind.io/wikitext/wikitext-2-v1.zip}} we used the following computing infrastructure: 5 GeForce RTX 2080 Ti gpu cards. Our codebase is based on Pytorch\footnote{\url{pytorch.org/get-started}} and is publicly available on Github.\footnote{\url{github.com/Noahs-ARK/groc}} 
\subsection{Model Configuration}
\label{sec:model_config}
The prefix network used by all output embedding methods is a vanilla recurrent neural network based on the implementation by \citet{merity2016pointer}\footnote{\url{github.com/salesforce/awd-lstm-lm}} with 2 layers and 1024 LSTM units, regularized with hidden unit dropout of $0.65$ along the lines of \citet{Grave17}. The maximum length of the relational and definitional forms from Wordnet is set to 3 and 10 without search based on our computational budget.\footnote{We expect that a larger  budget would generally allow to increase these limits and obtain even better results.}
The embedding size is set to 300 for \texttt{penn} and 256 for \texttt{wikitext2}. For optimization we use Adam with a learning rate of $0.001$,  initial weight uniformly sampled in the range~$[-0.05, 0.05]$, and a batch size of 20 for $\texttt{penn}$ and \texttt{wikitext2}.  
\begin{table}[htp]
  \centering \small
  \def\arraystretch{1.2}\tabcolsep=3.4pt    
  \begin{tabular}{l c c c}
    \toprule
    \textbf{Hyperparameter} & abbrev. &   range &    trials \\
    \toprule
     Output dropout  & $r$ & $\{0,0.1, \ldots, 1.0\}$ & 10 \\
     Linear correction& $\mathit{cor}$  & $\{32, 64, 128\}$ & 3\\
     Adaptive cutoffs & $\mathit{cut}$  & $\{253,721,118,226,$  & 6\\                            &                 & $424,334\}$ & \\
     Output net depth & $k$ & $\{0,1,2,3,4\}$ & 4 \\
     Output net activation & $\mathit{act}$ &  $\{relu,selu,tanh\}$ & 3 \\ 
    \bottomrule 
  \end{tabular}
  \caption{Hyperparameters, range of values, and, number of trials required to search them. Adaptive cutoffs are read as follows: e.g.~ for 253 the cutoff array contains $\big[0.2*n, 0.5*n, 0.3*n\big],n=|\mathcal{V}|$ words per bin.
  \label{tab:hyper_bounds}}
  \vspace{-5mm}
\end{table}
We clip the norm of the gradient to 0.1 and unroll the network for 35 steps. The learning rate is multiplied by 0.1 if the development loss does not decrease for 4 consecutive epochs and we perform early stopping if there is no improvement for 8 consecutive epochs. 

\subsection{Hyperparameter Optimization} \label{sec:lm_hyper}
For all methods, the hyperparameter selection of output embedding dropout rate ($r$), output network depth ($k$) and activation ($\mathit{act}$), linear ``correction'', and adaptive frequency cutoffs was conducted by grid search over specific range of values given in Table \ref{tab:hyper_bounds} on development data. Note that not all the hyperparameters apply to all methods, as can be seen in Table \ref{tab:hyper_optimal} where we report the optimal hyperparameter values for each of the methods.
For all the baselines we performed exhaustive grid search on both datasets, but for our method we performed grid search only on \texttt{penn} and searched manually on \texttt{wikitext-2} by selecting values of hyperparameters that were ranked high based on the grid search on \texttt{penn} to avoid the
increased cost that comes with training our method (see speed comparison in Appendix \ref{sec:training_speed}).  The total number of trials for all methods including our ablations were 204 and 67 respectively for \texttt{penn} and \texttt{wikitext-2} respectively. Note that the reduced number of trials is due to not performing exhaustive search for our method and its ablations as explained above. 
\begin{table}[htp]
  \centering \small
  \def\arraystretch{1.2}\tabcolsep=1.5pt    
  \begin{tabular}{l ccccc | ccccc }
    \toprule
      & \multicolumn{5}{c}{\texttt{penn}} & \multicolumn{5}{c}{\texttt{wikitext2}} \\%
    \textbf{Method} &  $r$ & $\mathit{cor}$ & $\mathit{cut}$ & $k$ & $\mathit{act}$ &  $r$ & $\mathit{cor}$ & $\mathit{cut}$ & $k$ & $\mathit{act}$\\
    \toprule
    Lookup table     & 0.1 & -- & -- & -- & -- & 0.2 & -- & -- & -- & --  \\
    Convolutional    & 0.1 & 128 & -- & -- &-- & 0.1 & 182 & --  & -- & -- \\ 
    Tied             & 0.0 & -- & -- & -- & -- & 0.0& -- & -- & -- & --  \\
    Bilinear         & 0.5 & -- & -- & -- &   -- & 0.4 & -- & -- & -- & --\\
    Deep residual    & 0.5 & -- & -- & 4 & selu & 0.6 & -- & -- & 1 & selu \\
    Adaptive         & 0.3 & -- & 2k7k & -- & -- & 0.2 & -- & 6k21k &-- & --   \\\midrule 
    GroC (ours)      & 0.2 & -- & -- & 0 & -- & 0.2 & -- & -- & 1 & relu  \\  
    $-$ relations     & 0.3 & -- & -- & 0 & -- & 0.3 & -- & -- & 1 & selu \\
    $-$ definitions   & 0.2 & -- & -- & 0 & -- & 0.3 & -- & -- & 2 & relu \\
    $-$ both         & 0.3 & -- & -- & 0 & -- & 0.3 & -- & -- & 1 & selu \\
    \bottomrule
  \end{tabular}
  \caption{Best hyperparameter values per method.}
  \label{tab:hyper_optimal}
\end{table} The number of trials per method can be derived by multiplying the non-zero columns per row with the number of trials required for each column. 

\subsection{Development Scores} \label{sup:dev_scores}
Table \ref{tab:lm_closed_dev} displays the development scores and number of parameters along with the test perplexities for our model and all the baseline output embedding methods for our main experiment. The development scores for the models of the ablation study and for the base models of the coverage experiment have already been given in Table \ref{tab:ablation} in the main paper (Section \ref{sec:analysis}). Overall, we can observe that in most cases the  ranking based on the development scores is indicative of the ranking of the methods according to the test scores.

\begin{table}[htp]
  \centering \small
  \def\arraystretch{1.2}\tabcolsep=3.5pt    
  \begin{tabular}{l rrr rrr }
    \toprule
      & \multicolumn{3}{c}{\texttt{penn}} & \multicolumn{3}{c}{\texttt{wikitext2}} \\ 
    \textbf{Method} & {$|\Theta|$ }  &  {dev.} & {test}  & {$|\Theta|$}  &  {dev.} & {test} \\
    \toprule
    Lookup table                                       & 13M & 93.5 & 90.8 & 23M & 113.8 & 108.3  \\
    Convolutional             & 13M & 104.0 & 101.6 & 23M & 121.2 & 116.6   \\ 
    Tied                      & 10M & 88.6 & 86.2 & 15M & 101.0 & 97.3 \\
    Bilinear                   & 10M & 87.0 & 83.7 & 15M & 101.3 & 95.9 \\
    Deep residual           & 10M & 84.0 & 80.5 & 15M & 100.1 & 94.7 \\
    Adaptive                &  8M & 84.0 & 79.3 & 9M  & 95.8 & 90.7  \\\midrule
    GroC (ours)               &  9M & {72.5} & {69.5} & 9M & {87.0}  & {82.5} \\ 
    \bottomrule
    
  \end{tabular}
  \caption{Development and test scores on conventional language modeling benchmarks with closed vocabulary. $|\Theta|$ denotes the total number of model parameters.  \label{tab:lm_closed_dev}}
  \vspace{-10mm}
\end{table}

\subsection{Training Speed} \label{sec:training_speed}
Table \ref{tab:train_speed} displays the average training speed per epoch in seconds for each of the methods. This experiment was run on a single, dedicated\footnote{By dedicated GPU card here we mean that no other processes were using the GPU card when we performed the experiments for each of the methods.} GeForce RTX 2080 Ti. As we mentioned in Section \ref{sec:compositional}, even though our model has vocabulary-size independent parameterization it is not independent of the computation that is required to encode the vocabulary. This has a negative impact on the training speed of \groc~, making it a bit slower than the Convolutional method, namely $1.3\times$ slower. 

To mitigate this problem we recommend training \groc~ with sparse updates for the output embedding parameters as described in the main paper (Section \ref{sec:training}). Concretely, at each training iteration with probability $p$ we make a full update and keep the output embedding frozen otherwise. The rest of the network is trained with full updates as before. 
 We can observe that this optimization strategy makes \groc~ nearly as efficient as the baselines with $p=0.1$ or $p=0.3$. In particular, it becomes even faster than the convolutional baseline by $2.1\times$. Furthermore, our best model with $p=0.3$ which is much faster reaches 75.3 perplexity on \texttt{penn} without additional hyperparameter optimization which is still $-4$ points lower than the second best, adaptive output embedding; tuning the model from scratch should likely lead to even better results. This is quite encouraging because it means that the benefits of our model need not come with a large computational cost. In future work, the training speed could be optimized even further by devising specialized efficient training methods for compositional outputs.  

\begin{table}[htp]
  \centering \small
  \def\arraystretch{1.2}\tabcolsep=9.5pt    
  \begin{tabular}{l ccccc ccccc }
    \toprule
     \textbf{Method} & \texttt{penn} & \texttt{wikitext-2}  \\
    \toprule
    Lookup table     &  19.5  & 59.5\\
    Convolutional    &  201.2 & 1301.9 \\ 
    Tied             &  18.6  & 53.6\\
    Bilinear         &  35.0  & 120.1\\
    Deep residual    &  61.2  & 114.5\\
    Adaptive         &  27.2  & 77.6\\\midrule 
    GroC (ours)      &  259.8 & 1813.5\\  
    $-$ 10\% updates &  236.3 & 1627.7\\
    $-$ 30\% updates &  173.5 & 1262.9\\
    $-$ 50\% updates &  131.5 & 936.4\\
    $-$ 70\% updates &  95.2  & 669.0\\
    $-$ 90\% updates &  46.0  & 299.0\\
    \bottomrule
  \end{tabular}
  \caption{Training speed for each method. We report the average time in seconds to complete one epoch.}
  \label{tab:train_speed}
\end{table}

\subsection{Comparison with State-of-the-Art Models} \label{sec:sota}

Table \ref{tab:PTB} displays several state-of-the-art models  which have number of parameters ranging from 9M to 20M on Penn Treebank. We can observe that our model which has only 9.7M parameters achieves better performance than all the models that have lower than or equal to 21M parameters and even the model by \citet{inan17} which has 24M parameters. Note that our model has lower perplexity than the pointer sentinel mixture model by \citet{merity2016pointer} and the neural cache model by \citet{Grave17} while having 11M less parameters than them. 

\begin{table}[ht]
	\footnotesize
	\centering
   \def\arraystretch{1.3}\tabcolsep=2.3pt  
	\begin{tabular}{lcc}
		\toprule
		\bf Model &   $|\Theta|$ &   test \\
		\midrule
		\citet{mikolov2012context} -- RNN-LDA  &  9M$^\ddagger$ & 92.0 \\ 
		\citet{zaremba2014recurrent} -- LSTM & 20M & 82.7 \\
		\citet{gal2016theoretically} -- Var. LSTM & 20M & 78.6 \\
		\citet{kim2016character} -- CharCNN & 19M & 78.9 \\
		\citet{merity2016pointer} -- Pointer Sentinel-LSTM & 21M &  70.9 \\
		\citet{grave2016improving} -- LSTM + cont. cache  & - & 72.1 \\
		\citet{inan17} -- Tied Variational LSTM  & 24M & 73.2 \\
		\citet{zilly2016recurrent} -- Variational RHN & 23M  & 65.4 \\
		\citet{zoph2016neural} -- NAS Cell & 25M & 64.0 \\
		\citet{merity2017regularizing} -- AWD-LSTM   & 24M & 58.8  \\
		 \midrule 
		Ours -- LSTM  & 10M &  86.2 \\ 
		Ours -- LSTM + \groc~ (sur,rel,def)  & 9.7M &  69.5  \\  
		\bottomrule 
    \end{tabular}
    \vspace{-2mm}
    \caption{Comparison with state-of-the-art models of comparable size to that of \citet{Grave17} and \citet{merity2016pointer} on the \texttt{penn} dataset.}
	\label{tab:PTB}
\end{table} 
\begin{figure*}
    \centering
    \includegraphics[height=4.5cm,width=0.493\textwidth]{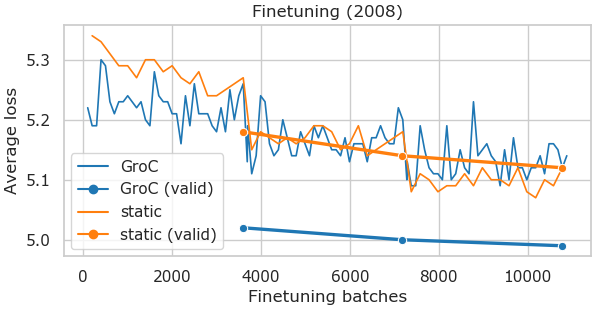}  \includegraphics[height=4.5cm,width=0.493\textwidth]{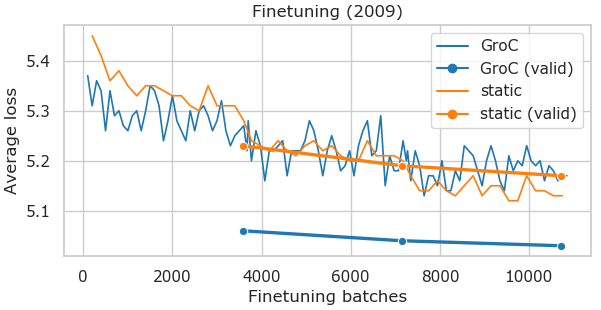}
    \includegraphics[height=4.5cm,width=0.493\textwidth]{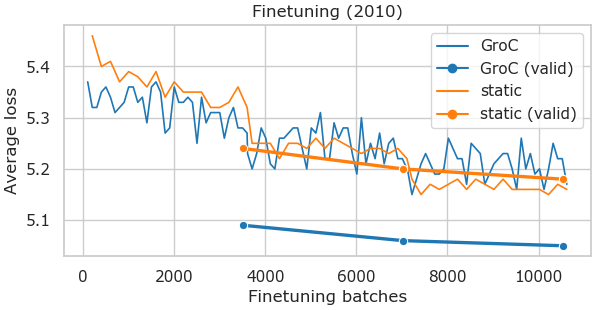}
    \includegraphics[height=4.5cm,width=0.493\textwidth]{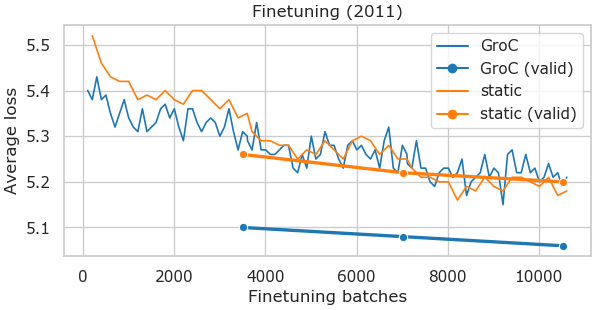}
    \caption{Training and validation loss for \groc~ and the tied model during finetuning on near domains.}
    \label{fig:near_plots}
\end{figure*}
Moreover, it is very close to the other models which have around 23-25M parameters without being highly regularized (weight dropout, input dropout) or having advanced optimization strategies (SGD + ASGD, finetuning) like AWD-LSTM \cite{merity2016pointer}. Training larger models and investigating the potential of competing with even higher capacity models is an interesting direction which we hope will be explored in future studies.

\section{Cross-Domain Language Modeling}
\label{sec:app_cross}
For the experiment in cross-domain language modeling, we used the following computing infrastructure: 2 GeForce RTX 2080 Ti and 2 TITAN RTX GPUs to train and finetune our \groc~ models, and 2 Tesla P100 GPUs to train and finetune the baselines and to perform hyperparameter search.

\subsection{Finetuning Dynamics}
\label{sec:app_cross_finetuning}

\begin{figure}[ht]
    \centering
    \includegraphics[height=4.5cm,width=0.485\textwidth]{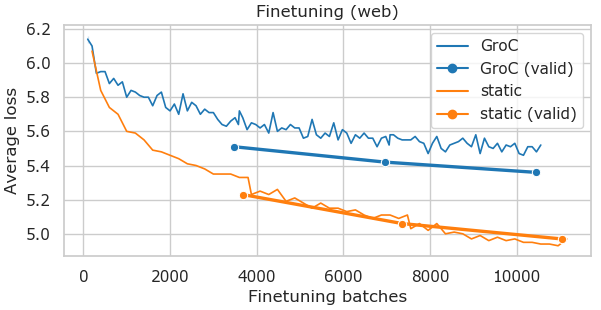}
    \includegraphics[height=4.5cm,width=0.485\textwidth]{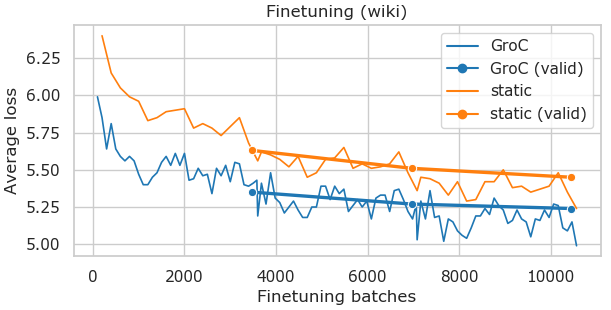}
    \caption{Training and validation loss for \groc~ and the tied model during finetuning, on far domains.}
    \label{fig:far_plots}
\end{figure}

Figures \ref{fig:near_plots} and \ref{fig:far_plots} show the loss on the training and validation data for the target domain during finetuning. 
\groc~ generalizes better from the training to the validation data than the tied model, consistently having lower validation loss. 
The training loss for \groc~ consistently starts out lower than that the of the tied model, showing that it has less difficulty adapting to the new data, and ends up higher, indicating greater regularization vs the tied model.

The \texttt{web} dataset is a clear outlier, in which the tied model improves much more dramatically than in any other domain. The difference in validation performance here is reflected in the test perplexity (Table \ref{tab:lm_cross}) but does not have a clear explanation.

\begin{table*}
  \centering \small
  \def\arraystretch{1.1}\tabcolsep=6pt    
  \begin{tabular}{p{0.1mm} l c  | c c c c | c  c }
    \toprule
    & & & \multicolumn{4}{c}{{{near} domains}} & \multicolumn{2}{c}{{{far} domains}} \\
    & \textbf{Model} & {\hspace{3mm}  \texttt{2007}} $\mathbf{\rightarrow}$     & {  \texttt{2008}}
      & { \texttt{2009}} &
       {  \texttt{2010}} & 
        {  \texttt{2011}} & 
        {   \texttt{Web}} &
      {   \texttt{Wiki}} \\ 
    \toprule
     &  Tied + finetuning    & -- & 167.44  &  175.95  & 177.46 & 180.63 & 144.13 & 232.06  \\
    \midrule
     &  Grounded + finetuning       &  -- &  146.84  & 152.29 & 155.27 & 158.21 & 212.99  & 188.25  \\
    \bottomrule 
  \end{tabular}
  \vspace{-2mm}
  \caption{Validation perplexity for finetuned models on cross-domain language modeling.}
  \label{tab:lm_cross_valid}
\end{table*}

\subsection{Data}
\label{sec:app_cross_data}

As described in Section \ref{sec:cross}, the choice of data and preprocessing used for the cross-domain experiments are based on \citet{Grave17}. News Crawl and Common Crawl can be downloaded from the WMT 2014 website.\footnote{\url{www.statmt.org/wmt14/translation-task.html}} WikiText-103 was downloaded from Salesforce website\footnote{\url{blog.einstein.ai/the-wikitext-long-term-dependency-language-modeling-dataset/}}. For the News Crawl datasets, the first 2M tokens of the English data for each year were used as the train set, the next 2M tokens as the validation set, and the next 10M tokens as the test set. The same procedure was used for \texttt{web} (Common Crawl), for which we used the English portion of the English-German aligned data. 
While \citet{Grave17} describes the Common Crawl data as shuffled at the sentence level, we found that most sentences seemed closely related to adjacent sentences, so after creating train/valid/test splits for this dataset we re-shuffled each file. WikiText-103 comes divided into train/valid/test splits, so we used the first 2M/2M/10M tokens of each split respectively for our dataset. All data was then tokenized using the Europarl tokenizer\footnote{\url{statmt.org/europarl/v7/tools.tgz}} and lowercased.

Our data preprocessing can be replicated with the script \texttt{create-data.sh}, available with the code for \groc~.\footnote{\url{github.com/<anon>/groc}}

\subsection{Finetuning Validation Results}
Because no target-domain training is required for most of our cross-domain experiments, validation scores were not computed for most model-domain combinations; however, we report the validation perplexity for the finetuned models in Table \ref{tab:lm_cross_valid}, to aid in replication.

\subsection{Hyperparameter Selection}
\label{sec:app_cross_hyper}

\begin{figure}[h]
    \centering
    \includegraphics[width=0.49\textwidth]{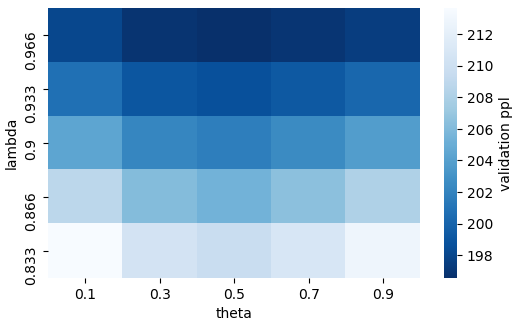}
    \caption{Validation accuracy for various hyperparameter settings on the \texttt{2008} validation set.}
    \label{fig:2008_search}
\end{figure}

Cache hyperparameters were selected via grid search, with $\theta$, the flattening hyperparameter described in \citet{grave2016improving}, ranging over 5 values from 0 to 1, and $\lambda$ ranging over 5 values from 0.833 to 0.966 (bounds which were selected based on the optimal hyperparameter ranges in \citep{grave2016improving}). Perplexity of a model trained on \texttt{2007} and evaluated on the \texttt{2008} validation set was the metric used to select the optimal hyperparameters: $\lambda=0.966$ for unigram and neural cache and $\theta=0.5$ for neural cache. Because the cache is only used during evaluation, this hyperparameter search was quite efficient to carry out using the tied model, requiring no additional training, only 25 evaluation runs on the validation set. This hyperparameter search is illustrated in Figure \ref{fig:2008_search}. 

\begin{figure}
    \centering
    \includegraphics[width=0.48\textwidth]{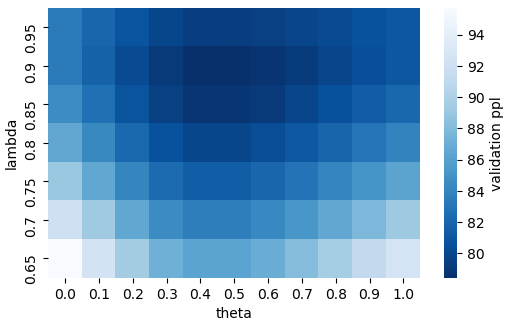}
     \caption{Validation accuracy for various hyperparameter settings on the \texttt{penn} validation set.}
    \label{fig:penn_search}
\end{figure}

\vspace{-1mm}
We then used the same hyperparameters for all cache models. This provides a slight advantage to the tied model, as the optimal hyperparameters for \groc~ might be different from those selected with the tied model.
A cache size of 5,000 was used during hyperparameter tuning, but at test time we used 10,000 for all experiments based on its use in \citetalias{}{Grave17}. Figure \ref{fig:penn_search} shows a separate hyperparameter search performed over the \texttt{penn} validation set to confirm the accuracy of our neural cache reimplementation. Compare to Figure 2a in \citet{grave2016improving}; note their $\lambda$ is 1 minus ours.

For \groc~, we also selected a downweighting hyperparameter $dw$, based on validation performance on the \texttt{wiki} dataset only. We searched over 5 values (0.1, 0.3, 0.5, 0.7, and 0.9) using \groc~ with the neural cache, and selected $dw=0.1$ as the best value with a validation ppl of 154.01.

\end{document}